\documentclass[sigconf,prologue,dvipsnames,table]{acmart}

\usepackage{amsmath,amsfonts}
\usepackage{natbib} 
\usepackage{algorithmic}
\usepackage{color}
\usepackage{graphicx}
\usepackage{enumitem}
\usepackage{multirow}
\usepackage{tabu}
\usepackage{tabularx}
\usepackage{textcomp}
\usepackage{xcolor}
\usepackage{svg}
\usepackage{caption}
\usepackage{subcaption}
\usepackage{ulem}

\usepackage{titlesec}
\titleformat{\paragraph}[runin]{\bfseries\normalsize}{\theparagraph}{1em}{}

\AtBeginDocument{%
  \providecommand\BibTeX{{%
    \normalfont B\kern-0.5em{\scshape i\kern-0.25em b}\kern-0.8em\TeX}}}

\copyrightyear{2022} 
\acmYear{2022} 
\setcopyright{rightsretained} 
\acmConference[JCDL '22]{The ACM/IEEE Joint Conference on Digital Libraries in 2022}{June 20--24, 2022}{Cologne, Germany}
\acmBooktitle{The ACM/IEEE Joint Conference on Digital Libraries in 2022 (JCDL '22), June 20--24, 2022, Cologne, Germany}
\acmDOI{10.1145/3529372.3530938}
\acmISBN{978-1-4503-9345-4/22/06}

\begin{document}

\title[X-SCITLDR: Cross-Lingual Extreme Summarization of Scholarly Documents]{%
X-SCITLDR: Cross-Lingual \texorpdfstring{\\}{}%
Extreme Summarization of Scholarly Documents%
}

\author{Sotaro Takeshita}
\affiliation{%
  \institution{%
  University of Mannheim}
  \city{Mannheim}
  \country{Germany}
}
\email{sotaro.takeshita@uni-mannheim.de}
\orcid{0000-0002-6510-7058}

\author{Tommaso Green}
\affiliation{%
  \institution{%
  University of Mannheim}
  \city{Mannheim}
  \country{Germany}
}
\email{tommaso.green@uni-mannheim.de}
\orcid{0000-0002-2231-2385}

\author{Niklas Friedrich}
\affiliation{%
  \institution{%
  University of Mannheim}
  \city{Mannheim}
  \country{Germany}
}
\email{nfriedri@mail.uni-mannheim.de}

\author{Kai Eckert}
\affiliation{%
  \institution{Hochschule der Medien}
  \city{Stuttgart}
  \country{Germany}
}
\email{eckert@hdm-stuttgart.de}

\author{Simone Paolo Ponzetto}
\affiliation{%
  \institution{%
  University of Mannheim}
  \city{Mannheim}
  \country{Germany}
}
\email{ponzetto@uni-mannheim.de}

\renewcommand{\shortauthors}{Takeshita et al.}

\begin{abstract}
The number of scientific publications nowadays is rapidly increasing, causing information overload for researchers and making it hard for scholars to keep up to date with current trends and lines of work. Consequently, recent work on applying text mining technologies for scholarly publications has investigated the application of automatic text summarization technologies, including extreme summarization, for this domain. However, previous work has concentrated only on monolingual settings, primarily in English. In this paper, we fill this research gap and present an abstractive cross-lingual summarization dataset for four different languages in the scholarly domain, which enables us to train and evaluate models that process English papers and generate summaries in German, Italian, Chinese and Japanese. We present our new X-SCITLDR dataset for multilingual summarization and thoroughly benchmark different models based on a state-of-the-art multilingual pre-trained model, including a two-stage `summarize and translate' approach and a direct cross-lingual model. We additionally explore the benefits of intermediate-stage training using English monolingual summarization and machine translation as intermediate tasks and analyze performance in zero- and few-shot scenarios.
\end{abstract}

\ccsdesc[500]{Computing methodologies~Natural language processing}
\ccsdesc[300]{Computing methodologies~Natural language generation}
\ccsdesc[300]{Computing methodologies~Language resources}

\keywords{Scholarly document processing, Summarization, Multilinguality}

\maketitle

\section{Introduction}

For years, the number of scholarly documents has been steadily increasing \cite{bornmann_growth_2015}, thus making it difficult for researchers to keep up to date with current publications, trends and lines of work. Because of this problem, approaches based on Natural Language Processing (NLP) have been developed to automatically organize research papers so that researchers can consume information in ways more efficient than just reading a large number of papers. For instance, citation recommendation systems provide a list of additional publications given an initial `seed' paper, in order to reduce the burden of literature reviewing \cite{bhagavatula_content-based_2018,medic_improved_2020}. One approach is to identify relevant sentences in the paper based on automatic classification \cite{jin_hierarchical_2018}. This approach to information distillation is taken further by fully automatic text summarization, where a long document is used as input to produce a shorter version of it covering essential points \cite{collins_supervised_2017,yasunaga_scisummnet_2019}, possibly a TLDR-like `extreme' summary \cite{cachola_tldr_2020}. Similar to the case of manually-created TLDRs, the function of these summaries is to help researchers quickly understand the main content of a paper without having to look at the full manuscript or even the abstract.

Just like in virtually all areas of NLP research, most successful approaches to summarization rely on neural techniques using supervision from labeled data. This includes neural models to summarize documents in general domains such as news articles \cite{see_get_2017,liu_text_2019}, including cross- and multi-lingual models and datasets \cite{scialom_mlsum_2020,varab_massivesumm_2021}, as well as specialized ones e.g., the biomedical domain \cite{mishra14}.

For the task of summarizing research papers, most available datasets are in English only, e.g., CSPubSum/CSPubSumExt \cite{collins_supervised_2017} and ScisummNet \cite{yasunaga_scisummnet_2019}, with community-driven shared tasks also having concentrated on English as \textit{de facto} the only language of interest \cite{jaidka2019clscisumm,chandrasekaran2019overview}. But while English is the main language in most of the research communities, especially those in the science and technology domain, this limits the accessibility of summarization technologies for the researchers who do not use English as the main language (e.g., many scholars in a variety of areas of humanities and social and political sciences). We accordingly focus on the problem of cross-lingual summarization of scientific articles -- i.e., produce summaries of research papers in languages different than the one of the original paper -- and benchmark the ability of state-of-the-art multilingual transformers to produce summaries for English research papers in different languages. Specifically, we propose the new task of \textbf{cross-lingual extreme summarization of scientific papers (CL-TLDR)}, since TLDR-like summaries have shown much promise in real-world applications such as search engines for academic publications like Semantic Scholar\footnote{\url{https://www.semanticscholar.org/product/tldr}}.

In order to evaluate the difficulty of CL-TLDR and provide a benchmark to foster further research on this task, we create a new multilingual dataset of TLDRs in a variety of different languages (i.e., German, Italian, Chinese, and Japanese). Our dataset consists of two main portions: a) a translated version of the original dataset from \citet{cachola_tldr_2020} in German, Italian and Chinese to enable comparability across languages on the basis of post-edited automatic translations; b) a dataset of human-generated TLDRs in Japanese from a community-based summarization platform to test performance on a second, comparable human-generated dataset. Our work complements seminal efforts from  \citet{fatima_novel_2021}, who compile an English-German cross-lingual dataset from the Spektrum der Wissenschaft / Scientific American and Wikipedia, in that we focus on extreme summarization, build a dataset of expert-derived multilingual TLDRs (as opposed to leads from Wikipedia), and provide additional languages.

The contributions of this work are as follows.
\begin{itemize}[leftmargin=3mm]
    \item We propose the \textbf{new task of cross-lingual extreme summarization of scientific articles} (CL-TLDR).
    \item We create \textbf{the first multilingual dataset for extreme summarization of scholarly papers} in four different languages.
    \item Using our dataset, \textbf{we benchmark the difficulty of cross-lingual extreme summarization} with different models built on top of state-of-the-art pre-trained language models \cite{lewis_bart_2020,liu_multilingual_2020}, namely a two-step approach based on monolingual summarization and translation, and a multilingual transformer model to directly generate summaries crosslingually.
    \item  Our dataset may not be sufficient to successfully fine-tune large pre-trained language models for direct cross-lingual summarization. Consequently, to tackle this data scarcity problem, \textbf{we additionally present experiments using intermediate fine-tuning techniques}, which have shown to be effective to improve performance of pre-trained multilingual language models on many downstream NLP tasks \cite[\textit{inter alia}]{gururangan_dont_2020,phang2019sentence,pruksachatkun-etal-2020-intermediate,glavas_xhate-999_2020}.
\end{itemize}
The remainder of this paper is organized as follows. We first summarize in Section \ref{sec:scitldr} seminal work on monolingual extreme summarization for English from \citet{cachola_tldr_2020}, on which our multilingual extension builds upon. We next introduce our new dataset for cross-lingual TLDR generation in Section \ref{sec:xscitldr}. We present our cross-lingual models and benchmarking experiments in Section \ref{sec:models} and \ref{sec:experiments}, respectively. Section \ref{sec:related} provides an overview of relevant previous work in monolingual and multilingual summarization. We wrap up our work with concluding remarks and directions for future work in Section \ref{sec:conclusion}.

\section{%
English Monolingual TLDR Summarization
}
\label{sec:scitldr}

\begin{table}
\centering
\begin{tabularx}{\columnwidth}{| X |}
\hline
\textbf{Abstract:} We propose a method for meta-learning reinforcement learning algorithms by searching over the space of computational graphs which compute the loss function for a value-based model-free RL agent to optimize. [...] \\
\textbf{Introduction: }Designing new deep reinforcement learning algorithms that can efficiently solve across a wide variety of problems generally requires a tremendous amount of manual effort. [...] \\
\textbf{Conclusion:} In this work, we have presented a method for learning reinforcement learning algorithms. We design\\a general language for representing algorithms which compute the loss function for [...]\\ 
\hline
\textbf{TLDR:} We meta-learn RL algorithms by evolving computational graphs which compute the loss function for a value-based model-free RL agent to optimize. \\
\hline
\end{tabularx}
\caption{An example of a TLDR summary for a research paper. Source: \url{https://openreview.net/forum?id=0XXpJ4OtjW}}
\label{ta:tldr_sample}
\end{table}

The main part of our new cross-lingual dataset consists of an automatically-translated, post-edited version of the SCITLDR dataset from \citet{cachola_tldr_2020}, who presented seminal work on the topic of extreme summarization of scientific publications for English.

SCITLDR is a dataset composed of pairs of research papers and corresponding summaries: in contrast to other existing datasets, this dataset is unique because of its focus on extreme summarization, i.e., very short, TLDR-like summaries, and consequently high compression ratios -- cf.\  the compression ratio of 238.1\% of SCITLDR vs.\ 36.5\% of CLPubSum \cite{collins_supervised_2017}. An example of a TLDR summary is presented in Table~\ref{ta:tldr_sample}, where we see how information from different summary-relevant sections of the paper (typically, in the abstract, introduction and conclusions) is often merged to provide a very short summary that is meant to help readers quickly understand the key message and contribution of the paper.

The original SCITLDR dataset consists of 5,411 TLDRs for 3,229 scientific papers in the computer science domain: it is divided into a training set of  1,992 papers, each with a single gold-standard TLDR, and  dev and test sets of 619 and 618 papers each, with 1,452 and 1,967 TLDRs, respectively (thus being multi-target in that a document can have multiple gold-standard TLDRs). The summaries consist of TLDRs written by authors and collected from the peer review platform OpenReview\footnote{\url{https://openreview.net}}, as well as human-generated summaries from peer-review comments found on the same platform. But while this dataset encourages the development of scholarly paper summarization systems, the original version only supports English as the target language, even though research across different fields can be conducted by researchers who use various languages from all over the world. Therefore, in this work, we extend the summaries in SCITLDR to support four additional languages, namely German, Italian, Chinese and Japanese.

\section{X-SCITLDR}
\label{sec:xscitldr}

In this section, we describe the creation of the X-SCITLDR dataset and briefly present some statistics to provide a quantitative overview. Our dataset is composed of two main sources:

\begin{itemize}[leftmargin=3mm]
  \item a manually post-edited version of the original SCITLDR dataset \cite{cachola_tldr_2020} for German, Italian and Chinese (X-SCITLDR-PostEdit).
  \item a human-generated dataset of expert-authored TLDRs harvested from a community-based summarization platform for Japanese (X-SCITLDR-Human).
\end{itemize}

\begin{table*}[ht]
    \centering
    \begin{tabu} to \textwidth {| X[1.5] | X[6] |}
        \multicolumn{2}{c}{a) German}\\
        \hline
        Original Summary       & The \textbf{paper} presents a multi-view framework for improving sentence representation in NLP tasks using generative and discriminative objective architectures.                   \\\hline
        Automatic Translation & \colorbox{BurntOrange}{Das Papier} präsentiert einen Multi-View-Rahmen zur Verbesserung der Satzrepräsentation in NLP-Aufgaben \ldots %
        \\\hline
        Postedited  Version & \colorbox{BurntOrange}{Der Artikel} präsentiert einen Multi-View-Rahmen zur Verbesserung der Satzrepräsentation in NLP-Aufgaben \ldots %
        \\\hline
        \multicolumn{1}{c}{}\\
        \multicolumn{2}{c}{b) Italian}\\\hline
        Original Summary       & The paper provides a full characterization of permutation invariant and equivariant linear layers for \textbf{graph data}.  \\\hline
        Automatic Translation &         L'articolo fornisce una caratterizzazione completa degli strati lineari invarianti di permutazione ed equivarianti per i \colorbox{BurntOrange}{dati del grafico}.
        \\\hline
        Postedited Version & L'articolo fornisce una caratterizzazione completa dei layer lineari invarianti o equivarianti per la permutazione per i \colorbox{BurntOrange}{dati del grafo}.
        \\\hline

    \end{tabu}
    \caption{%
    Example of a post-editing correction (wrong sense): `\textsf{Papier}' means a generic piece of paper but not a research paper in German (`\textsf{Artikel}'). Similarly, English `graph' needs to be translated as `\textsf{grafo}' as opposed to `\textsf{grafico}' (English: `diagram').
    }
    \label{ta:postedit_1}
\end{table*}
\begin{table*}[ht]
    \centering
    \begin{tabu} to \textwidth {| X[1.5] | X[6] |}
        \multicolumn{2}{c}{a) German}\\
        \hline
        Original Text       & The paper proposes a framework for constructing spherical \textbf{convolutional networks} based on a novel synthesis of several existing concepts.                                \\\hline
        Automatic Translation & Das Papier schlägt einen Rahmen für die Konstruktion von sphärischen \colorbox{BurntOrange}{Faltungsnetzen} vor, der auf einer neuartigen Synthese mehrerer bestehender Konzepte beruht. \\\hline
        Postedited Version & Die Arbeit schlägt einen Rahmen für die Konstruktion von sphärischen \colorbox{BurntOrange}{Convolutional Networks vor}, der auf einer neuartigen Synthese mehrerer bestehender Konzepte beruht. \\\hline
        \multicolumn{1}{c}{}\\
        \multicolumn{2}{c}{b) Italian}\\\hline
        Original Text & We present a novel iterative algorithm based on generalized low rank models for computing and interpreting \textbf{word embedding models}.
        \\\hline
        Automatic Translation & Presentiamo un nuovo algoritmo iterativo basato su modelli generalizzati di basso rango per il calcolo e l'interpretazione dei \colorbox{BurntOrange}{modelli di incorporazione delle parole}.
        \\\hline
        Postedited Version & Presentiamo un nuovo algoritmo iterativo basato su modelli generalizzati di basso rango per il calcolo e l'interpretazione dei \colorbox{BurntOrange}{modelli di word embedding}.\\\hline
        
    \end{tabu}
    \caption{Example of a post-editing correction (terminological English-preserving translation). `Convolutional network' can be translated in German as `\textsf{faltendes Netz}' or `\textsf{Faltungsnetz}', whereas `word embedding' can be translated as both `\textsf{incorporazione}' or `\textsf{immersione delle parole}' in Italian. We reduce variability in summaries by keeping the English domain-specific term in the target-language summaries.}
    \label{ta:postedit_2}
\end{table*}

\paragraph{X-SCITLDR-PostEdit.} Given the overall quality of automatic translators \cite{daniele19}, we opt for a hybrid machine-human translation process of post-editing \cite{2013-post-editing} in which human annotators correct machine-generated translations as post-processing to achieve higher quality than when only using an automatic system. Although current machine translation systems arguably provide nowadays high-quality translations, a manual correction process is still necessary for our data, especially given their domain specificity. In Tables \ref{ta:postedit_1} and \ref{ta:postedit_2}, we present examples of how translations are corrected by human annotators, and the reasons for the correction. These can be grouped into two cases:
\begin{itemize}[leftmargin=3mm]
    \item[a)] Wrong translation due to selected wrong sense (Table \ref{ta:postedit_1}). In this case, the machine translation system has problems selecting the domain-specific sense and translation of the source term.
    \item[b)] Translation of technical terms (Table \ref{ta:postedit_2}). To avoid having the same technical term being translated in different ways, we reduce the sparsity of the translated summaries and simplify the translation task by preserving technical terms in English.
\end{itemize}
Both cases indicate the problems of the translation system with domain-specific terminology. For the underlying translation system, we use DeepL\footnote{\url{https://www.deepl.com/translator}}. After the automatic translation process, we asked graduate students in computer science courses who are native speakers in the target language to fix incorrect translations.

\begin{table*}
\centering
\begin{tabular}{|l|r|r|r|r||r|r|r|r|} 
\hline
& \multicolumn{4}{c||}{Documents} & \multicolumn{4}{c|}{Summaries}\\
\cline{2-9}
& \multicolumn{1}{l|}{\begin{tabular}[c]{@{}l@{}}\# documents\\(train/dev/test)\end{tabular}} & \multicolumn{1}{l|}{\begin{tabular}[c]{@{}l@{}}\# words\end{tabular}} &  \multicolumn{1}{l|}{\begin{tabular}[c]{@{}l@{}}vocabulary\\size\end{tabular}} & \multicolumn{1}{l||}{\begin{tabular}[c]{@{}l@{}}average\\\# words\\per doc\end{tabular}} & \multicolumn{1}{l|}{\begin{tabular}[c]{@{}l@{}}\# words\end{tabular}} & \multicolumn{1}{l|}{\begin{tabular}[c]{@{}l@{}}vocabulary\\size\end{tabular}} & \multicolumn{1}{l|}{\begin{tabular}[c]{@{}l@{}}average\\\# words\\per summary\end{tabular}} & \multicolumn{1}{l|}{\begin{tabular}[c]{@{}l@{}}compression\\ratio (\%)\end{tabular}} \\ 
\hline

EN & \multirow{4}{*}{1,992/619/618} & \multirow{4}{*}{370,244} & \multirow{4}{*}{20,819} & \multirow{4}{*}{5,000} & 47,574 &  6,725 & 23.88 & 244.57 \\ %
\cline{1-1}\cline{6-9}
DE & & & & & 43,929 & 13,808 & 22.05 & 264.87 \\ 
\cline{1-1}\cline{6-9}
IT & & & & & 48,050 & 7,127 & 24.12 & 242.14 \\ 
\cline{1-1}\cline{6-9}
ZH & & & & & 47,711 & 7,953 & 23.95 & 243.86 \\ 

\hline\hline
JA & 1,606/199/199 & 306,815 & 14,769 & 10,000 & 121,989 & 6,706 & 75.91 & 131.73\\ %
\hline
\end{tabular}
\caption{Statistics of our dataset (X-SCITLDR).}
\label{ta:stats}
\end{table*}

\paragraph{X-SCITLDR-Human.} We complement the translated portion of the original TLDR dataset with a new dataset in Japanese crawled from the Web. For this, we collect TLDRs of scientific papers from a community-based summarization platform, arXivTimes\footnote{\url{https://arxivtimes.herokuapp.com}}.
This Japanese online platform is actively updated by users who voluntarily add links to papers and a corresponding user-provided short summary.
The posted papers cover a wide range of machine learning related topics (e.g., computer vision, natural language processing and reinforcement learning). This second dataset portion allows us to test with a dataset for extreme summarization of research papers in an additional language and, crucially, with data entirely written by humans, which might result in a writing style different from the one in X-SCITLDR-PostEdit. That is, we can use these data not only to test the capabilities of multilingual summarization in yet another language but, more importantly, test how much our models are potentially overfitting by too closely optimizing to learn the style of the X-SCITLDR-PostEdit summaries.

\vspace{1em} \noindent
In Table \ref{ta:stats}, we present various statistics of our X-SCITLDR dataset for both documents and summaries from the original English (EN) SCITLDR data\footnote{Slight differences with respect to the statistics from \citet[Table 1]{cachola_tldr_2020}, e.g., different average number of words per summary (21 vs.\ 23.88), are due to a different tokenization (we use SpaCy: \url{https://spacy.io}).} and our new dataset in four target languages\footnote{Vocabulary sizes are computed after lemmatization with SpaCy.}. \mbox{SCITLDR} and \mbox{X-SCITLDR-PostEdit} (DE/IT/ZH) have a comparably high compression ratio (namely, the average number of words per document to the average number of words per summary) across all four languages, thus indeed requiring extreme cross-lingual compression capabilities. While summaries in German, Italian and Chinese keep the compression ratio close to the original dataset in English, summaries in the Japanese dataset come from a different source and consequently exhibit rather different characteristics, most notably longer documents and summaries. Manual inspection reveals that Japanese documents come from a broader set of venues than SCITLDR, since arXivTimes includes many ArXiv, ACL and OpenReview manuscripts (in contrast to SCITLDR, whose papers overwhelmingly come from ICLR, cf.\ \cite[Table 9]{cachola_tldr_2020}), whereas Japanese summaries often contain more than one sentence. Despite having both longer documents and summaries, the Japanese data still exhibit a very high compression ratio (cf.\ datasets for summarization of both scientific and non-scientific documents having typically a compression ratio <40\%), which indicates their suitability for evaluating extreme summarization in the scholarly domain.
\section{Models for Cross-lingual TLDR}
\label{sec:models}

We next present a variety of models that we use to benchmark the feasibility and difficulty of the task of cross-lingual extreme summarization of scientific papers (henceforth: CL-TLDR). Our cross-lingual models are able to automatically generate summaries in a target language given abstracts in English. For this, we build upon the original work from \cite{cachola_tldr_2020} and focus on \textit{abstractive} summarization, since this has been shown to outperform \textit{extractive summarization} in a variety of settings.

\paragraph{BART / mBART.} In our experiments, we use BART \cite{lewis_bart_2020} and its multilingual variant mBART \cite{liu_multilingual_2020} as underlying summarization models. They are both transformer-based \cite{vaswani_attention_2017} pre-trained generative language models, which are trained with an objective to reconstruct noised text in an unsupervised sequence-to-sequence fashion. While BART only uses an English corpus for pre-training, mBART learns from a corpus containing multiple languages. These pre-trained BART/mBART models can be further trained (i.e., fine-tuned) in order to be applied to downstream tasks of interest like, for instance, summarization, translation or dialogue generation. We use BART/mBART as our underlying models, since these have been shown in previous work to perform well on the task of extreme summarization \cite{lewis_bart_2020}. We follow \citet{ladhak-etal-2020-wikilingua} and use BART/mBART as components of two different architectures, namely: a) two-step approach to cross-lingual summarization, i.e., summarization via BART and translation using Machine Translation (MT) (Section \ref{sec:mt}); b) a direct cross-lingual summarization system obtained by fine-tuning mBART with input articles from English, and summaries from the target language (Section \ref{sec:direct}).

\begin{figure}[tb]
    \centering
    \includegraphics[width=.85\columnwidth]{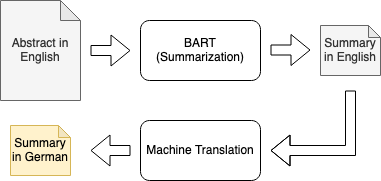}
    \caption{Overview of our two-stage `summarize and translate' approach (Section \ref{sec:mt}). It first summarizes an English abstract, then translates the generated summary into the target language.}
    \label{fig:pipeline}
\end{figure}

\subsection{Two-stage Cross-lingual Summarization: Summarize and Translate}
\label{sec:mt}

A first solution to the CL-TLDR task is to combine a monolingual summarization model with machine translation in the target language (we call this approach \textbf{EnSum-MT}). This approach is composed of two stages. The model first takes an English text as input and generates a summary in English: the English summary is then automatically translated into the target language using machine translation (Figure \ref{fig:pipeline}).\footnote{Like for the creation of the multilingual portion of our dataset, we opt again for DeepL for all our languages (cf.\ Section \ref{sec:xscitldr}).} This model does not rely on any cross-lingual signal: it merely consists of two independent modules for translation and summarization and does not require any cross-lingual dataset to train the summarization model.
While this system is conceptually simple, such a pipeline approach is known to cause an error propagation problem \cite{zhu_ncls_2019}, since errors of the first stage (i.e., summarization) get amplified in the second stage (i.e., translation) leading to overall performance degradation.

\begin{figure*}[tb]
    \centering
    \includegraphics[width=\textwidth]{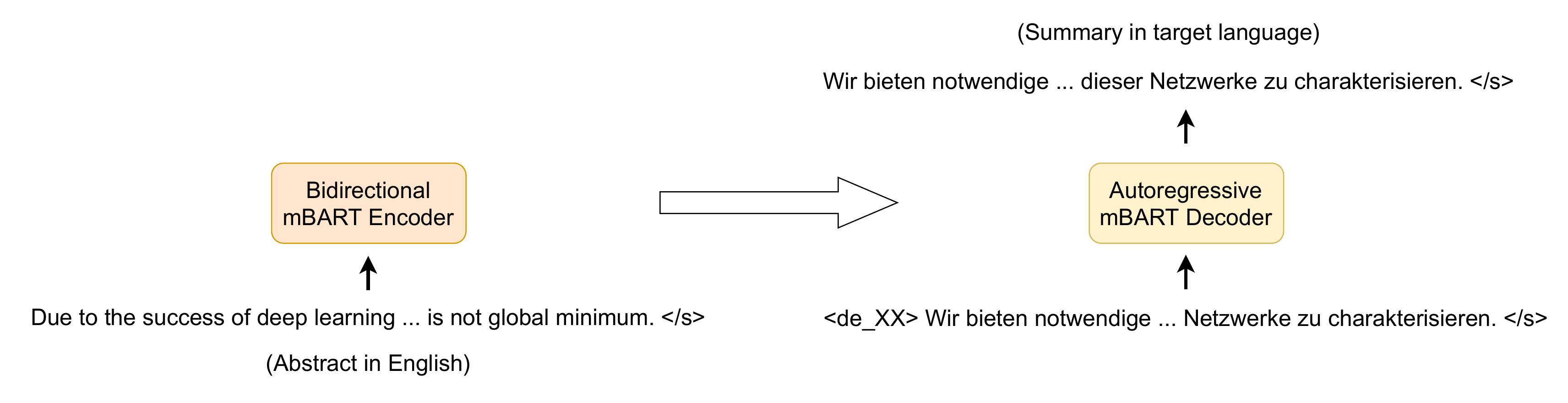}
    \caption{mBART learns to take an English abstract and generate a summary in the target language (here, German). We can control the target language by providing a language token (\textless de\textgreater in the figure).}
    \label{fig:mbart_enc_dec}
\end{figure*}

\subsection{Direct Cross-lingual Summarization}
\label{sec:direct}

A second approach to CL-TLDR is to directly perform cross-lingual summarization using a pre-trained multilingual language model (we call this method \textbf{CLSum}). For this, we investigate the use of a pre-trained multilingual denoising autoencoder, namely mBART \cite{liu_multilingual_2020}, and use cross-lingual training data from our new X-SCITLDR dataset to fine-tune mBART and generate summaries in the target languages given abstracts in English, as depicted in Figure \ref{fig:mbart_enc_dec}. We follow \citet{liu_multilingual_2020} and control the target language by providing a language token to the decoder.

\begin{figure}[t]
    \centering
    \includegraphics[width=\columnwidth]{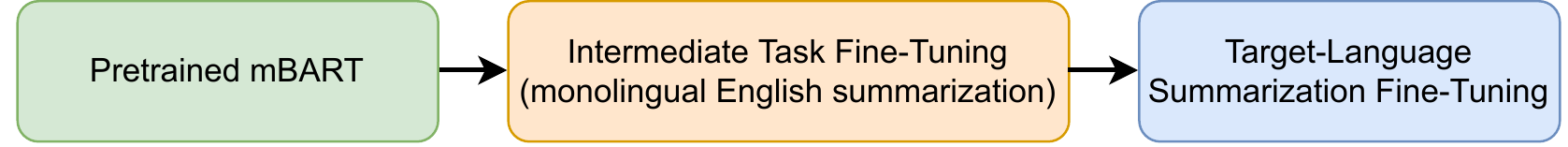}
    \caption{Intermediate task fine-tuning (CLSum+EnSum): We insert an additional training stage between pre-training and cross-lingual summarization fine-tuning in which the pre-trained language model learns to summarize monolingually in English.}
    \label{fig:task_ip}
\end{figure}

\begin{figure}[t]
    \centering
    \includegraphics[width=\columnwidth]{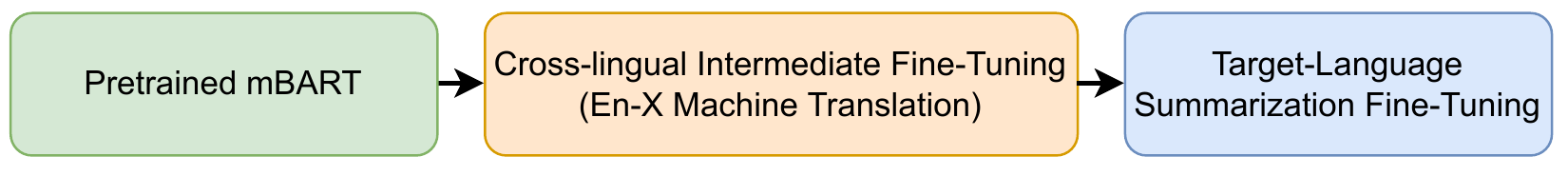}
    \caption{Cross-lingual intermediate fine-tuning (CLSum+ MT). We insert an additional training stage between pre-training and downstream cross-lingual summarization by fine-tuning the pre-trained language model to translate from English into the target language.}
    \label{fig:lang_ip}
\end{figure}

\paragraph{Intermediate task and cross-lingual fine-tuning.} Our training dataset is relatively small compared to datasets for general domain summarization \cite{narayan_dont_2018,hermann_teaching_2015}.
To mitigate this data scarcity problem, we investigate the effectiveness of intermediate fine-tuning, which has been reported to improve a wide range of downstream NLP tasks (see, among others, \cite{gururangan_dont_2020,phang2019sentence,pruksachatkun-etal-2020-intermediate,glavas_xhate-999_2020}). \citet{gururangan_dont_2020}, for instance, show that training pre-trained language models on texts in a domain/task similar to the target domain/task can boost the performance on the downstream task by injecting additional related knowledge into the models. Based on this observation, in our experiments, we investigate two strategies for intermediate fine-tuning: intermediate task and cross-lingual fine-tuning. 

\begin{itemize}[leftmargin=3mm]
    \item \textbf{Intermediate task fine-tuning (CLSum+EnSum).} We explore the benefits of using \textit{additional summarization data} other than the summaries in the target language and augment the training dataset with English data, i.e., the original SCITLDR data. That is, before fine-tuning on summaries in the target language (e.g., German), we train the model on English TLDR summarization as auxiliary monolingual summarization task to provide additional summarization capabilities (Figure \ref{fig:task_ip}).
    \item \textbf{Cross-lingual intermediate fine-tuning (CLSum+MT).} Direct cross-lingual summarization requires the model to learn both translation and summarization skills, arguably a difficult task given our small dataset.\footnote{In preliminary experiments mBART often failed to generate in the target language after fine-tuning it on our cross-lingual dataset, thus confirming the need to augment with translation data (see also previous findings from \citet{ladhak-etal-2020-wikilingua}).} To alleviate this problem, we investigate a model that is trained on machine translation, before being fine-tuned on the summarization task. For this, we automatically translate the English abstracts into the target language and use these synthetic data as training data for fine-tuning on the task of automatically translating abstracts (Figure \ref{fig:lang_ip}).
\end{itemize}

\section{Experiments}
\label{sec:experiments}

\paragraph{Input documents.} We follow \citet{cachola_tldr_2020} and rely in all our experiments on an input consisting of abstracts only, since they showed that it yields similar results when compared to using the abstract, introduction, and conclusion sections together. Even more importantly, using only abstracts enables the applicability of our models also to those cases where only the abstracts are freely available and we do not have open access to the complete manuscripts. The average length of an abstract is 185.87 words for X-SCITLDR-PostEdit (EN/DE/IT/ZH) with an average compression ratio of 12.64\%, and 190.92 words for X-SCITLDR-Human (Japanese) and a compression ratio of 39.76\%.

\paragraph{Evaluation metrics.} We compute performance using a standard metric to automatically evaluate summarization systems, namely ROUGE \cite{lin-hovy-2003-automatic}. In the case of the post-edited portion of the X-SCITLDR dataset (X-SCITLDR-PostEdit, Section \ref{sec:xscitldr}), the gold standard can contain multiple reference summaries for a given paper and abstract. Consequently, for Italian, German and Chinese we calculate ROUGE scores in two ways (\textit{avg} and \textit{max}) to account for these multiple references \cite{cachola_tldr_2020}. For \textit{avg}, we compute ROUGE F1 scores with respect to the different references and take the average score, whereas for \textit{max} we select the highest scoring one. The Japanese dataset does not contain multiple reference TLDRs: hence, we compute standard ROUGE F1 only. We test for statistical significance using sample level Wilcoxon signed-rank test with $\alpha = 0.05$ \cite{dror-etal-2018-hitchhikers}.

\paragraph{Hyperparameter tuning.} To find the best hyperparameters for each model, we use the development data and run a grid search using ROUGE-1 avg as a reference metric. We run experiments with learning rate $\in \{ 1 \cdot 10^{-5}, 3 \cdot 10^{-5}\}$ and random seed $\in \{1122, 22\}$ during fine-tuning, number of beams for beam search $\in \{2, 3\}$ and repetition penalty rate $\in \{0.8, 1.0\}$ during decoding. For all settings, we set the batch size equal to 1 and perform 8 steps of gradient accumulation. We use the AdamW optimizer \cite{loshchilov2019decoupled} with weight decay of 0.01 for 5 epochs without warm-up.

\paragraph{Training strategy.} To prevent the model from losing the knowledge acquired in pre-training during fine-tuning (i.e., catastrophic forgetting \cite{KemkerMAHK18}), we freeze the parameters of the embedding and decoder layers during intermediate task and cross-lingual fine-tuning \cite{maurya_zmbart_2021}, while we update the entire model during the final fine-tuning for downstream CL-TLDR.
Since mBART requires large memory space when we update the entire model, we utilize the DeepSpeed library\footnote{\url{https://www.deepspeed.ai}} to meet our infrastructure requirements.

\paragraph{Research questions.} We organize the presentation and discussion of our results around the following research questions:

\begin{itemize}[leftmargin=3mm]
    \item \textbf{RQ1}: Which \textit{architecture} -- i.e., two-stage or direct crosslingual summarization (Sections \ref{sec:mt} and \ref{sec:direct}) -- is best suited for the CL-TLDR task? How do the results compare for \textit{different kinds of cross-lingual data}, i.e., different portions of our dataset (X-SCITLDR-PostEdit vs.\ -Human, Section \ref{sec:xscitldr})?
    \item \textbf{RQ2}: Does intermediate-stage fine-tuning help improve direct CL-TLDR summarization?
    \item \textbf{RQ3}: How much data do we need to perform cross-lingual TLDR summarization? What is the performance in a zero-shot or few-shot setting?
\end{itemize}

\begin{table*}
\centering
\begin{tabular}{clllllll}
\multicolumn{1}{l}{Lang} & Model & \multicolumn{1}{l}{R1 (avg)} & \multicolumn{1}{l}{R2 (avg)} & \multicolumn{1}{l}{RL (avg)} & \multicolumn{1}{l}{R1 (max)} & \multicolumn{1}{l}{R2 (max)} & \multicolumn{1}{l}{RL (max)} \\ 
\hline
\hline
\multirow{4}{*}{DE} & EnSum-MT & \textbf{19.29} & \textbf{5.46} & \textbf{16.02} & \textbf{30.74} & \textbf{13.37} & \textbf{26.61} \\ 
\cline{2-8}
 & CLSum & 17.99 & 3.58 & 14.69 & 27.44 & 8.54 & 23.05 \\
 & CLSum+EnSum & 18.06 & 3.61 & 14.75 & 27.36 & 8.47 & 23.04 \\
 & CLSum+MT & 18.47 & 4.16 & 15.25 & 28.84 & 9.91 & 24.37 \\ 
\hline
\multirow{4}{*}{IT} & EnSum-MT & 20.76 & 6.88 & 17.46 & 31.53 & 14.96 & 27.51 \\ 
\cline{2-8}
 & CLSum & 21.20 & 6.15 & 17.54 & 30.98 & 12.77 & 26.25 \\
 & CLSum+EnSum & 20.47 & 6.14 & 17.39 & 30.13 & 12.61 & 26.32 \\
 & CLSum+MT & \textbf{21.71}$^\dagger$ & \textbf{7.04} & \textbf{18.11}$^\dagger$ & \textbf{32.34} & \textbf{14.44} & \textbf{27.76} \\ 
\hline
\multirow{4}{*}{ZH} & EnSum-MT & \textbf{27.06} & \textbf{8.69} & \textbf{23.26} & \textbf{40.41} & \textbf{18.18} & \textbf{35.39} \\ 
\cline{2-8}
 & CLSum & 23.03 & 5.76 & 20.27 & 34.11 & 11.77 & 30.12 \\
 & CLSum+EnSum & 22.62 & 5.52 & 19.88 & 33.42 & 11.43 & 29.45 \\
 & CLSum+MT & 23.28 & 5.97 & 20.27 & 35.15 & 12.54 & 30.72 \\
\hline
\hline
\end{tabular}
\caption{Results on the X-SCITLDR-PostEdit portion of our cross-lingual TLDR dataset (ROUGE-1,-2 and -L): English to German, Italian or Chinese TLDR-like summarization using post-edited, automatically-translated summaries of the English data from \citet{cachola_tldr_2020}. Best results per language and metric are bolded. Statistically significant improvements of the cross-lingual models (CLSum/+EnSum/+MT) with respect to the `summarize and translate' pipeline (EnSum-MT) are marked with $^\dagger$.%
}
\label{ta:result-post}
\end{table*}

\begin{table}
\centering
\begin{tabular}{llll}
Model & \multicolumn{1}{l}{Rouge-1} & \multicolumn{1}{l}{Rouge-2} & \multicolumn{1}{l}{Rouge-L} \\ 
\hline
\hline
EnSum-MT & 24.38 & 4.42 & 16.54 \\
\hline
CLSum & 30.94 & 4.66$^*$ & 20.34\\
CLSum+EnSum & \textbf{32.30} & \textbf{5.66} & \textbf{20.89} \\
CLSum+MT & \textbf{32.30} & 5.47 & 20.85 \\ 
\hline
\hline
\end{tabular}
\caption{Results on the X-SCITLDR-Human portion of our cross-lingual TLDR dataset (Rouge-1,-2 and -L): English to Japanese TLDR-like summarization using human-generated summaries from ArXivTimes. Best results per metric are bolded. Statistically non-significant improvements of the cross-lingual models (CLSum/+EnSum/+MT) with respect to EnSum-MT are marked with an asterisk ($^*$).}
\label{ta:result-human}
\end{table}

\paragraph{RQ1: Summarize and translate vs.\ direct cross-lingual-TLDR.}

We present overall results on the two main portions of our dataset, i.e., post-edited translations and human-generated summaries, in Table \ref{ta:result-post}
and \ref{ta:result-human}, respectively.

When comparing our two main architectures, namely our MT-based `summarize and translate' system (Section \ref{sec:mt}, EnSum-MT) vs.\ our multilingual encoder-decoder architecture based on mBART (Section \ref{sec:direct}, CLSum/+EnSum/+MT) we see major performance differences across the two dataset portions. While the MT-based summarization (EnSum-MT) is superior or comparable with all translated/post-edited TLDRs in German, Italian and Chinese (Table \ref{ta:result-post}), the direct cross-lingual summarization model (CLSum) improves results on native Japanese summaries by a large margin (Table \ref{ta:result-human}).

The differences in performance figures between X-SCITLDR-PostEdit (German, Italian and Chinese) and X-SCITLDR-Human (Japanese) are due to the different nature of the multilingual data, and how they were created. Post-edited data like those in German, Italian and Chinese are indeed automatically translated, and tend to better align to the also automatically translated English summaries, as provided as output of the EnSum-MT system. That is, since both summaries -- the post-edited reference ones and those automatically generated and translated -- go through the same process of automatic machine translation, they naturally tend to have a higher lexical overlap, i.e. a higher overlap in terms of shared word sequences. This, in turn, receives a higher reward from ROUGE, since this metric relies on n-gram overlap between system and reference summaries.
While two-stage cross-lingual summarization seems to better align with post-edited reference TLDRs, for human-generated Japanese summaries, we observe an opposite behavior. Japanese summaries indeed have a different style than those in English (and their post-edited multilingual versions from X-SCITLDR-PostEdit) and accordingly have a lower degree of lexical overlap with translated English summaries from EnSum-MT.

\begin{table}
\centering
\begin{tabularx}{\columnwidth}{| X |}
\multicolumn{1}{c}{a) gold standard}\\
\hline
\textbf{Abstract:}~ Convolution acts as a local feature extractor in convolutional neural networks (CNNs). However, the convolution operation is not applicable when the input data is supported on an irregular graph such as with social networks, citation networks, or knowledge graphs. This paper proposes the topology adaptive graph convolutional network (TAGCN), a novel graph convolutional network that generalizes CNN architectures to graph-structured data and provides a systematic way to design a set of fixed-size learnable filters to perform convolutions on graphs. [...] \\ 
\hline
\textbf{TLDR: }The paper introduces Topology Adaptive GCN to generalize convolutional networks to graph-structured data.\\
\hline
\textbf{German TLDR: } Die Arbeit führt Topology Adaptive GCN ein, um Convolutional Networks auf graph-strukturierte Daten zu verallgemeinern.\\
\hline
\multicolumn{1}{c}{}\\
\multicolumn{1}{c}{b) automatically generated summaries}\\
\hline
\textbf{EnSum-MT: }In diesem Beitrag wird das topologieadaptive graphische Faltungsnetzwerk (TAGCN) vorgeschlagen, das CNN-Architekturen auf graphisch strukturierte Daten verallgemeinert und einen systematischen Weg zur Entwicklung einer Reihe von lernfähigen Filtern fester Größe zur Durchführung von Faltungen auf Graphen bietet. \\ 
\hline
\textbf{CLSum: }Wir schlagen das topologie adaptive graph convolutional network (TAGCN) vor, ein neuartiges graphisches Convolutional Network, das CNN-Architekturen auf graphenstrukturierte Daten verallgemeinert. \\
\hline
\end{tabularx}
\caption{Example of gold-standard summaries and automatically generated versions.}
\label{ta:example_tldr}
\end{table}

\begin{table}
\centering
\begin{tabular}{lrr}
Language & \multicolumn{1}{l}{EnSum-MT} & \multicolumn{1}{l}{CLSum}\\ 
\hline
\hline
DE & 23.48 & 22.94 \\
IT & 24.17 & 22.73 \\
ZH & 25.90 & 19.76 \\
JA & 30.50 & 56.76 \\
\hline
\hline
\end{tabular}
\caption{Average summary length (number of tokens).}
\label{ta:avg-length}
\end{table}

To better understand the behavior of the system in light of the different performance on post-edited vs.\ human-generated data, we manually inspected the output of the two systems. Table \ref{ta:example_tldr} shows an example of automatically generated summaries for a given input abstract: it highlights that summaries generated using our cross-lingual models (CLSum) tend to be shorter and consequently `abstracter' than those created by translating English summaries (EnSum-MT). This, in turn, can hurt the performance of the cross-lingual models more in that, while we follow standard practices and used ROUGE F1, this metric has been found unable to address the problems with ROUGE recall, which rewards longer summaries, in the ranges of typical summary lengths produced by neural systems \cite{sun-etal-2019-compare}. Table \ref{ta:avg-length} presents the average summary lengths in different languages for our MT-based and cross-lingual systems: the numbers show that for languages in the X-SCITLDR-PostEdit portion of our dataset summaries are indeed shorter. Japanese cross-lingually generated summaries are instead longer, due to the reference summaries being comparably longer than those in SCITLDR (cf.\ Table \ref{ta:stats}, column 8 and 9).

\begin{table}
\centering
\begin{tabular}{lrrr}
Language & \multicolumn{1}{c}{Train} & \multicolumn{1}{c}{Val} & \multicolumn{1}{c}{Test} \\ 
\hline \hline
DE & 0.95 & 0.92 & 0.92 \\
IT & 0.79 & 0.78 & 0.78 \\
ZH & 0.96 & 0.95 & 0.94 \\
\hline \hline
\end{tabular}
\caption{Word-level Jaccard coefficients between automatically translated summaries and their post-edited versions.}
\label{ta:jaccard}
\end{table}

Within the post-edited portion of our dataset, EnSum-MT performs significantly better than the cross-lingual models in German and Chinese; however, there is generally no significant difference with cross-lingual models in Italian, where CLSum+MT is even able to achieve statistically significant improvements on average Rouge-1 and Rouge-L. To better understand such different behavior across  languages, we computed for each language the word-level Jaccard coefficients between the automatically translated summaries and their post-edited versions. As Table \ref{ta:jaccard} shows, the Italian post-edited translations contain much more edits than the other two languages. This, in turn, seems to disadvantage the two-stage pipelined model, whose output aligns more with the `vanilla' automatic translations.

We notice also differences in absolute numbers between German, Italian and Chinese, which could be due to the distribution of training data used to train the multilingual transformer \cite{liu_multilingual_2020}, with mBART being trained on more Chinese than Italian data. However, German performs worst among the three languages, despite mBART being trained on more German data than Chinese or Italian. Manual inspection reveals that German summaries tend to be penalized more because of differences in word compounding between reference and generated summaries: while there exist proposals to address this problem in terms of language-specific pre-processing \cite{Frefel20}, we opt here for a standard evaluation setting equal for all languages. Moreover, German summaries tend to contain less English terms than, for instance, Italian summaries (6.78 vs.\ 4.88 English terms per summary on average in the test data), which seems to put the cross-lingual model at an advantage (cf.\ English terminology in EnSum-MT vs.\ CL-Sum in Table \ref{ta:example_tldr}). The performance gap between EnSum-MT and CLSum is the largest on the Chinese dataset, which shows that it is more challenging for mBART to learn to summarize from English into a more distant language \cite{lauscher_zero_2020}.

\paragraph{RQ2: The potential benefits of intermediate fine-tuning.} In the second set of experiments, we compare the performance of our `base' cross-lingual model with those using intermediate-stage training for learning the summarization and translation tasks from additional data. Specifically, we compare target-language fine-tuning of mBART (CLSum) with additional intermediate task fine-tuning on English monolingual summarization (CLSum+EnSum) and cross-lingual intermediate fine-tuning using machine translation on synthetic data (CLSum+MT). The rationale behind these experiments is that in the direct cross-lingual setting the model needs to acquire both summarization and translation capabilities, which requires a large amount of cross-lingual training data, and thus might be hindered by the limited size of our dataset.

Including additional training on summarization based on English data (CLSum+EnSum) has virtually no effects on the translated portion of SCITLDR (Table \ref{ta:result-post}) for German, and even degrades performance on Italian and Chinese. This is likely because English TLDR summaries are well aligned with their post-edited translations and virtually bring no additional signal while requiring the decoder to additionally translate into an additional language (i.e., English and the target language). On the contrary, CLSum+EnSum improves on all ROUGE metrics for Japanese (Table \ref{ta:result-human}): this is because, as previously mentioned, the ArXivTimes data have a different style from SCITLDR and thus English TLDRs provide an additional training signal that help to improve results for the summarization task.

Including MT-based pre-training, i.e., fine-tuning mBART on machine translation from English into the target language, and then on cross-lingual summarization (CLSum+MT) improves over simple direct cross-lingual summarization (CLSum) on all languages -- a finding in line with results from \citet{ladhak-etal-2020-wikilingua} for WikiHow summarization. This highlights the importance of fine-tuning the encoder-decoder for translation before actual fine-tuning for the specific cross-lingual task, thus injecting general translation capabilities into the model.

\begin{table}
\centering
\begin{tabular}{clrrr}
Lang & Model & \multicolumn{1}{c}{R1 (avg)} & \multicolumn{1}{c}{R2 (avg)} & \multicolumn{1}{c}{RL (avg)} \\ 
\hline
\hline
\multirow{3}{*}{DE} & CLSum & 2.67 & 0.46 & 2.58 \\
\cline{2-5}
 & CLSum+EnSum & 3.46 & 0.70 & 3.32 \\
 & CLSum+MT & 14.42 & 2.04 & 10.75 \\ 
\hline
\multirow{3}{*}{IT} & CLSum & 4.83 & 0.97 & 4.41 \\
\cline{2-5}
 & CLSum+EnSum & 5.87 & 1.29 & 5.35 \\
 & CLSum+MT & 16.11 & 3.48 & 12.38 \\ 
\hline
\multirow{3}{*}{ZH} & CLSum & 0.64 & 0.06 & 0.61 \\
\cline{2-5}
 & CLSum+EnSum & 0.79 & 0.10 & 0.76 \\
 & CLSum+MT & 17.88 & 3.60 & 13.95 \\
\hline
\multirow{3}{*}{JA} & CLSum & 2.34 & 0.59 & 2.06 \\
\cline{2-5}
 & CLSum+EnSum & 2.37 & 0.68 & 2.17 \\
 & CLSum+MT & 29.43 & 4.29 & 18.27 \\ 

\hline
\hline
\end{tabular}
\caption{Performance in zero-shot settings.  no intermediate fine-tuning (CLSum) vs.\ intermediate task (+EnSum) and cross-lingual fine-tuning (+MT).}
\label{ta:results_zeroshot}
\end{table}

\begin{figure*}[t]
     \centering
     \begin{subfigure}[b]{0.3\textwidth}
         \centering
         \includegraphics[width=\textwidth]{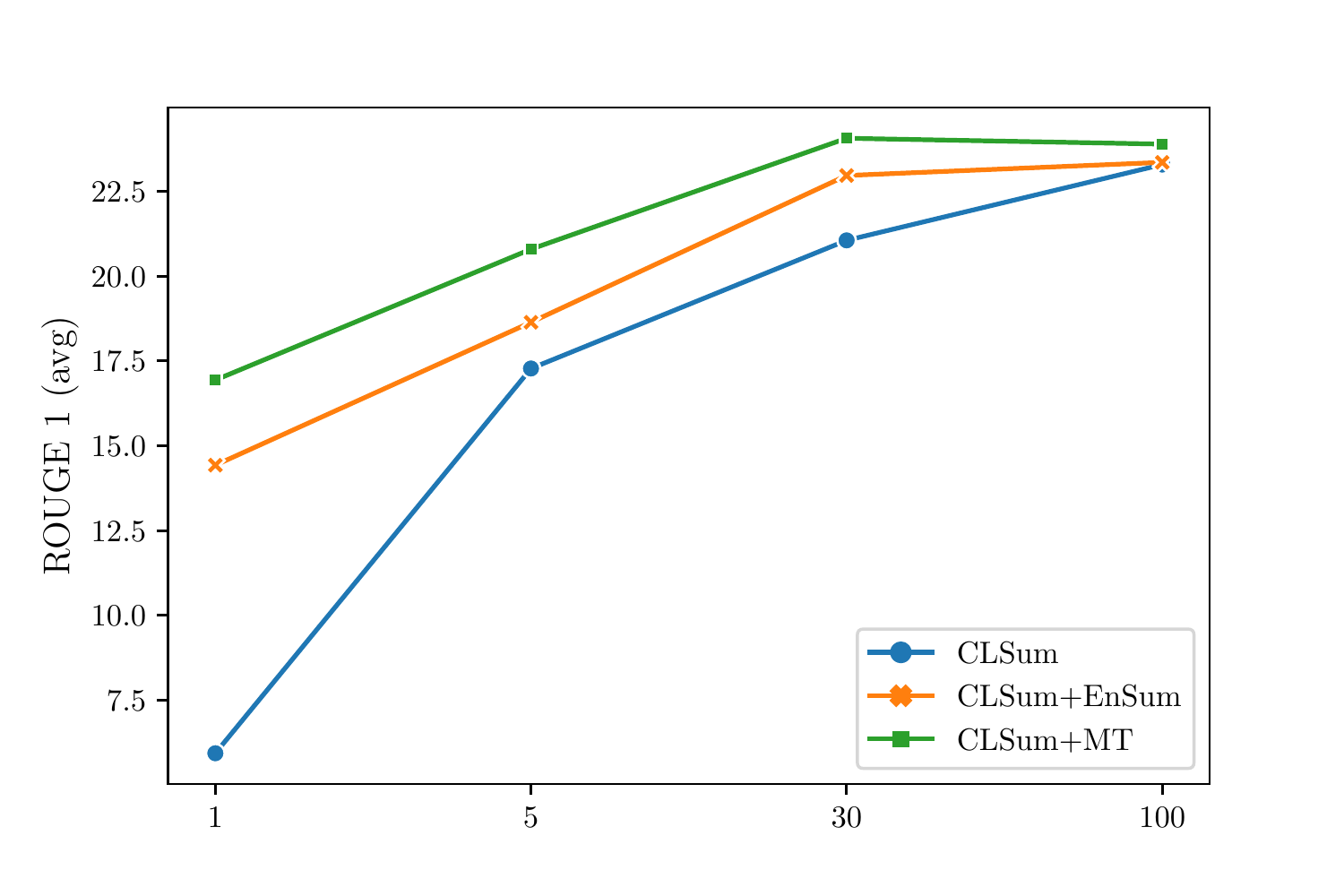}
         \caption{Rouge-1 (avg)}
         \label{fig:R1}
     \end{subfigure}
     \hfill
     \begin{subfigure}[b]{0.3\textwidth}
         \centering
         \includegraphics[width=\textwidth]{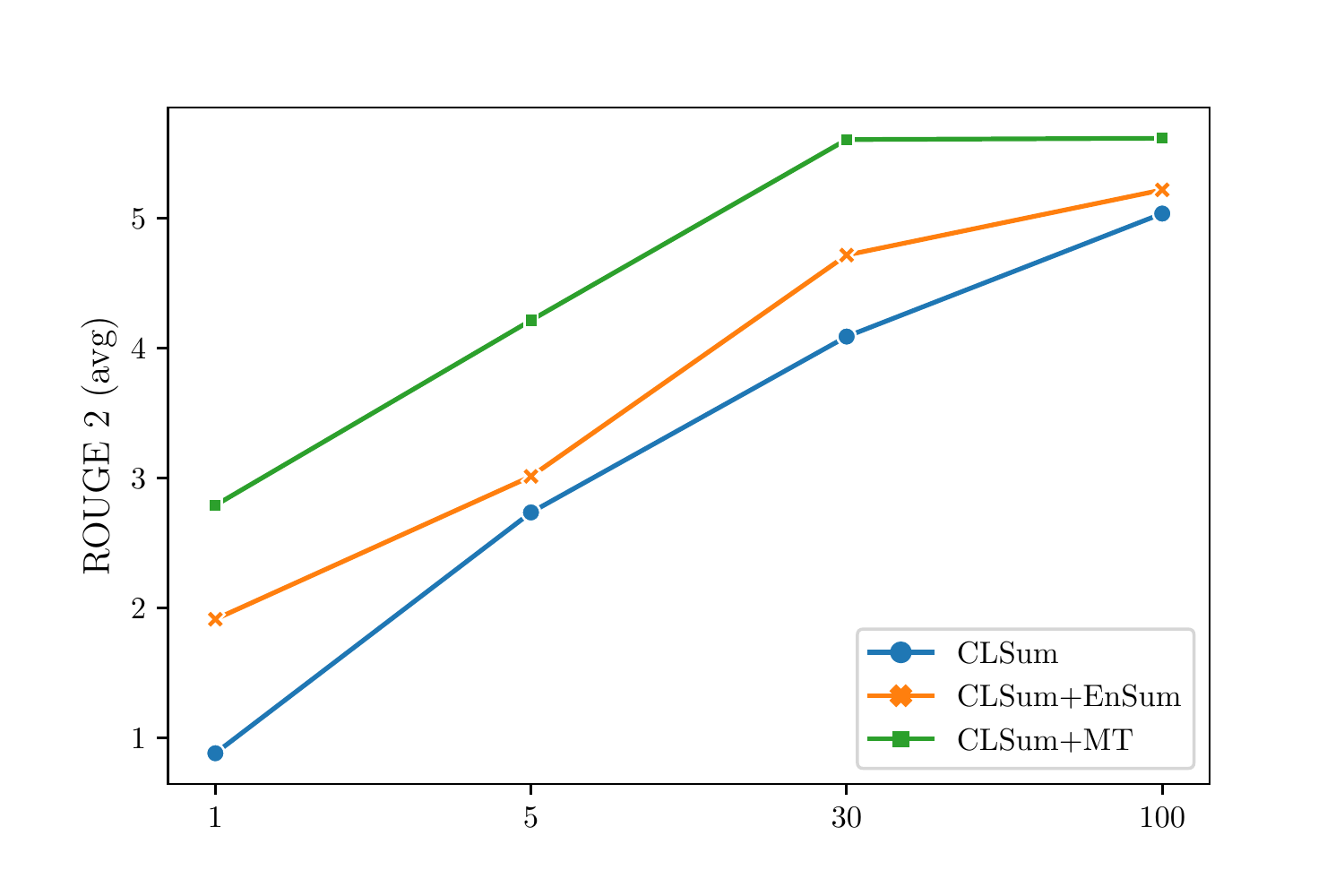}
         \caption{Rouge-2 (avg)}
         \label{fig:R2}
     \end{subfigure}
     \hfill
     \begin{subfigure}[b]{0.3\textwidth}
         \centering
         \includegraphics[width=\textwidth]{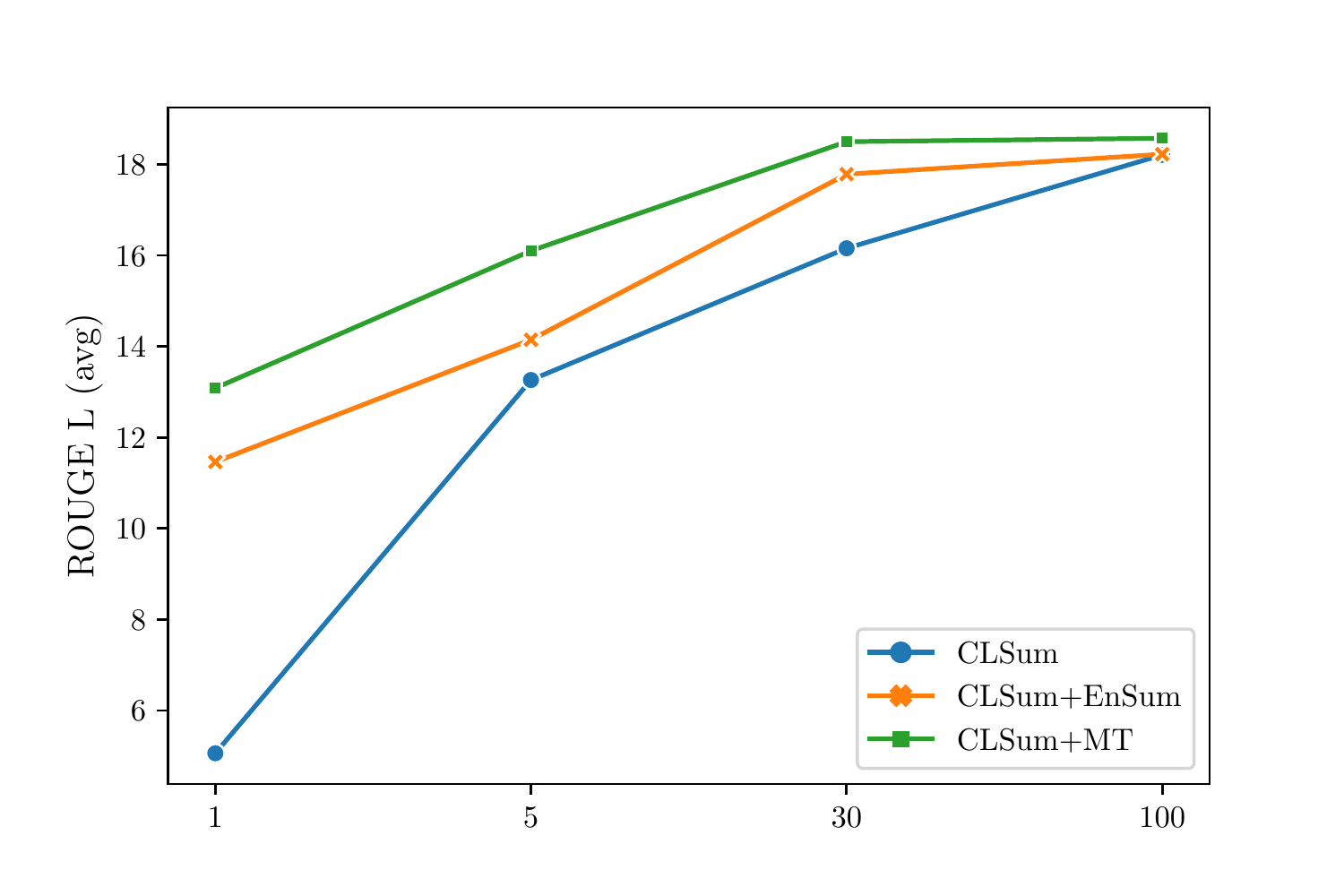}
         \caption{Rouge-L (avg)}
         \label{fig:RL}
     \end{subfigure}
        \caption{Few-shot results without (CLSum) and with intermediate task (+EnSum) and cross-lingual fine-tuning (+MT) for different sizes of the training data in the target language (i.e., different number of \textit{shots}): 1\%, 5\%, 30\% and 100\% of the X-SCITLDR training set of each target language.}
        \label{fig:ROUGE_scores}
\end{figure*}

\begin{table*}
\centering
\begin{tabular}{clrrrrrrrrrrrrrr}
\multicolumn{1}{l}{} &  & \multicolumn{4}{c}{Rouge-1 (avg)} & \multicolumn{1}{l}{} & \multicolumn{4}{c}{Rouge-2 (avg)} & \multicolumn{1}{l}{} & \multicolumn{4}{c}{Rouge-L (avg)} \\ 
\cline{3-6}\cline{8-11}\cline{13-16}
Lang & Model & 1\% & 5\% & 30\% & 100\% & \multicolumn{1}{l}{} & 1\% & 5\% & 30\% & 100\% & \multicolumn{1}{l}{} & 1\% & 5\% & 30\% & 100\% \\ 
\hline
\hline
\multirow{3}{*}{DE} & CLSum & 11.30 & 14.67 & 13.74 & 17.99 &  & 1.30 & 1.94 & 2.43 & 3.58 &  & 9.33 & 11.78 & 11.25 & 14.69 \\
\cline{2-16}
 & CLSum+EnSum & 15.01 & 13.72 & 17.92 & 18.03 &  & 1.65 & 1.82 & 2.94 & 3.56 &  & 12.24 & 10.85 & 14.24 & 14.74 \\
 & CLSum+MT & 9.30 & 16.01 & 18.42 & 18.28 &  & 1.19 & 2.83 & 3.89 & 3.99 &  & 7.90 & 13.04 & 15.07 & 15.07 \\ 
\hline
\multirow{3}{*}{IT} & CLSum & 9.51 & 17.18 & 18.25 & 21.20 &  & 1.50 & 3.17 & 4.27 & 6.15 &  & 8.20 & 13.75 & 15.20 & 17.54 \\
\cline{2-16}
 & CLSum+EnSum & 16.36 & 18.06 & 21.03 & 20.47 &  & 2.46 & 3.35 & 5.92 & 6.14 &  & 13.12 & 14.22 & 17.51 & 17.39 \\
 & CLSum+MT & 15.40 & 18.28 & 21.56 & 21.71 &  & 2.82 & 4.61 & 6.80 & 7.04 &  & 12.27 & 15.23 & 17.73 & 18.11 \\ 
\hline
\multirow{3}{*}{ZH} & CLSum & 0.76 & 9.97 & 20.45 & 23.03 &  & 0.09 & 1.13 & 4.30 & 5.76 &  & 0.73 & 8.77 & 17.55 & 20.27 \\
\cline{2-16}
 & CLSum+EnSum & 4.47 & 13.18 & 23.06 & 22.62 &  & 0.31 & 2.03 & 5.57 & 5.52 &  & 4.23 & 11.55 & 19.84 & 19.88 \\
     & CLSum+MT & 14.80 & 18.11 & 24.05 & 23.28 & \multicolumn{1}{l}{} & 2.84 & 3.83 & 6.20 & 5.97 & \multicolumn{1}{l}{} & 12.84 & 15.53 & 20.63 & 20.27 \\
\hline
\multirow{3}{*}{JA} & CLSum & 2.17 & 27.29 & 31.78 & 30.94 &  & 0.63 & 4.70 & 5.36 & 4.66 &  & 1.98 & 18.75 & 20.63 & 20.34 \\
\cline{2-16}
 & CLSum+EnSum & 21.87 & 29.60 & 29.86 & 32.30 &  & 3.23 & 4.85 & 4.44 & 5.66 &  & 16.26 & 19.97 & 19.55 & 20.89 \\
 & CLSum+MT & 28.25 & 30.78 & 32.23 & 32.30 &  & 4.31 & 5.59 & 5.54 & 5.47 &  & 19.33 & 20.61 & 20.57 & 20.85 \\ 
\hline
\hline
\end{tabular}
\caption{Detailed few-shot performance on downstream CL-TLDR for each language and cross-lingual model with different percentages of X-SCITLDR training data for each target language.}
\label{ta:results_fewshot}
\end{table*}

\paragraph{RQ3: Zero- and few-shot experiments.} To better understand the contribution of intermediate fine-tuning and to analyze performance in the absence of multilingual summarization training data (i.e., in zero-shot settings), we present experiments in which we compare: a) using mBART with no fine-tuning (CLSum); b) fine-tuning mBART on English SCITLDR data only and evaluate performance on X-SCITLDR in our four languages; c) fine-tuning mBART on synthetic translations of abstracts only and testing on X-SCITLDR. These experiments are meant to quantify the \textit{zero-shot cross-lingual transfer} capabilities of the cross-lingual models (i.e., can we train on English summarization data only without the need of a multilingual dataset?) as well as to explore how much we can get away with summarization data at all (i.e., what is the performance of a system that is trained to simply translate abstracts?).

We present our results in Table \ref{ta:results_zeroshot}. The performance figures indicate that the zero-shot cross-lingual transfer performance of CLSum+EnSum is extremely low for all our four languages, with reference performance on English TLDRs from \citet{cachola_tldr_2020} being 31.1/10.7/24.4 R1/R2/RL (cf.\ Table 10, BART `abstract-only'), and barely improves over no fine tuning at all (CLSum). This suggests that robust cross-lingual transfer in our summarization task is more difficult than in other language understanding tasks (see for example the much higher average performance on the XTREME tasks \cite{hu2020xtreme}). The overall very good performance of CLSum+MT seems to suggest that robust cross-lingual summarization performance can still be achieved without multilingual summarization data through the `shortcut' of fine-tuning a multilingual pre-trained model to translate English abstracts, since these can indeed be seen as summaries (albeit of a longer length than our TLDR-like summaries) and can thus be merely translated so as to provide a strong baseline.

We finally present results for our models in a few-shot scenario to investigate performance using cross-lingual data with a limited number of example summaries (\textit{shots}) in the target language. Figure \ref{fig:ROUGE_scores} shows few-shot results averaged across all four target languages (detailed per-language results are given in Table \ref{ta:results_fewshot}) for different sizes of the training data from the X-SCITLDR dataset. The results highlight that, while CL-TLDR is a difficult task with the models having little cross-lingual transfer capabilities (as shown in the zero-shot experiments), performance can be substantially improved when combining a small amount of cross-lingual data, i.e. as little as 1\% of examples for each target language, and intermediate training. As the amount of cross-lingual training data increases, the benefits of intermediate fine-tuning become smaller and results for all models tend to converge. This indicates the benefits of intermediate fine-tuning in the scenario of limited training data such as, e.g., low-resource languages other than those in our resource. Our few-shot results indicate that we can potentially generate TLDRs for a multitude of languages by creating little labeled data in those languages, and at the same time leverage via intermediate fine-tuning labeled resources for English summarization and machine translation (for which there exist plenty of resources).

\section{Related Work}
\label{sec:related}

\paragraph{General-domain summarization datasets.} News article platforms play a major role when collecting data for summarization \cite{sandhaus_evan_new_2008,hermann_teaching_2015}, since article headlines provide ground-truth summaries. Narayan et al. \cite{narayan_dont_2018} propose a news domain summarization dataset with highly compressed summaries to provide a more challenging summarization task (i.e., extreme summarization). \citet{sotudeh_tldr9_2021} propose TLDR9+, another extreme summarization dataset that was collected automatically from a social network service.

\paragraph{Cross-lingual summarization datasets.} While there are growing numbers of cross-lingual datasets for natural language understanding tasks \cite{conneau_xnli_2018,ruder_xtreme-r_2021,liang_xglue_2020}, few datasets for cross-lingual summarization are available.  \citet{zhu_ncls_2019} propose to use machine translation to extend English news summarization to Chinese. To ensure dataset quality, they adopt round-trip translation by translating the original summary into the target language and back-translating the result to the original language for comparison, keeping the ones that meet a predefined similarity threshold. \citet{ouyang_robust_2019} create cross-lingual summarization datasets by using machine translation for low-resource languages such as Somali, and show that they can generate better summaries in other languages by using noisy English input documents with English reference summaries. Our work differs from these prior attempts in that our automatically translated summaries are corrected by human annotators, as opposed to providing silver standards in the form of automatic translations without any human correction. Recently,  \citet{ladhak-etal-2020-wikilingua} presented a large-scale multilingual dataset for the evaluation of cross-lingual abstractive summarization systems that is built out of parallel data from \mbox{WikiHow}. Even though it is a large high-quality resource of parallel data for cross-lingual summarization, this corpus is built from how-to guides: our dataset focuses instead on scholarly documents. Besides cross-lingual corpora, there are also large-scale multilingual summarization datasets for the news domain \cite{varab_massivesumm_2021,scialom_mlsum_2020}. The work we present here differs in that we focus on extreme summarization for the scholarly domain and we look specifically at the problem of \textit{cross-lingual} summarization in which source and target language differ.

\paragraph{Datasets for summarization in the scholarly domain.}
There are only a few existing summarization datasets for the scholarly domain and most of them are in English. SCITLDR \cite{cachola_tldr_2020}, the basis for our work on multilingual summarization, presents a dataset for research papers (see Section \ref{sec:scitldr} for more details). \citet{collins_supervised_2017} use author-provided summaries to construct
an extractive summarization dataset from computer science papers, with over 10,000 documents. \citet{cohan_discourse-aware_2018} regard abstract sections in papers as summaries and create large-scale datasets from two open-access repositories (arXiv and PubMed). \citet{yasunaga_scisummnet_2019} efficiently create a dataset for the computational linguistics domain by manually exploiting the structure of papers. \citet{meng_bringing_2021} present a dataset which contains four summaries from different aspects for each paper, which makes it possible to provide summaries depending on requests by users. \citet{lu_multi-xscience_2020} is a large-scale dataset for multi-document summarization for scientific papers, for which models need to summarize multiple documents.

The work closest to ours has been recently presented by \citet{fatima_novel_2021}, who introduce an English-German cross-lingual summarization dataset collected from German scientific magazines and Wikipedia. This resource is complementary to ours in many different aspects. While both datasets are in the scientific domain, their data includes either articles from the popular science magazine Scientific American / Spektrum der Wissenschaft or articles from the Wikipedia Science Portal. In contrast, out dataset includes scientific publications written by researchers for a scientific audience. Second, our dataset focuses on extreme, TLDR-like summarization, which we argue is more effective in helping researchers browse through many potentially relevant publications in search engines for scholarly documents. Finally, our summaries are expert-generated, as opposed to relying on the `wisdom of the crowds' from Wikipedia, and are available in three additional languages.

\paragraph{Summarization of scientific documents.} In recent years, there has been much work on the problem of summarizing scientific publications, a task that belongs to the wider area of scholarly data and text mining \cite{SaggionR17,beltagy-etal-2021-overview}. Scholarly document processing has gained much traction lately, due to the ever growing need to efficiently access large amounts of published information, e.g., in the COVID-19 pandemic \cite{nlp-covid19-2020-nlp,Esteva2021}. Research efforts in summarization have been arguably catalyzed by community-driven evaluation campaigns such as the CL-SciSumm shared tasks \cite{jaidka2019clscisumm,chandrasekaran2019overview}. Previous work on summarization has focused on specific features of scientific documents such as using citation contexts \cite{CohanG18,ZervaNNA20} or document structure \cite{ConroyD18,cohan-etal-2018-discourse}. Complementary to these efforts is a recent line of work on automatically generating visual summaries or graphical abstracts \cite{YangLKJOWH19,YamamotoLPGM21}. In our work, we build upon recent contributions on using multilingual pre-trained language models for cross-lingual summarization \cite{ladhak-etal-2020-wikilingua} and extreme summarization for English \cite{cachola_tldr_2020}, and bring these two lines of research together to propose the new task of cross-lingual extreme summarization of scientific documents.
\section{Conclusion}
\label{sec:conclusion}

In this paper, we presented X-SCITLDR, the first dataset for cross-lingual summarization of scientific papers. Our new dataset makes it possible to train and evaluate NLP models that can generate summaries in German, Italian, Chinese and Japanese from input papers in English. We used our dataset to investigate the performance of different architectures based on multilingual transformers, including a two-stage `summarize and translate' approach and a direct cross-lingual model. We additionally explored the potential benefits of intermediate task and cross-lingual fine-tuning and analyzed the performance in zero- and few-shot scenarios. For future work, we plan to investigate how to apply additional techniques designed for cross-lingual text generation such as training with multiple decoders \cite{zhu_ncls_2019}, automatically complementing our multilingual TLDRs with visual summaries \cite{YamamotoLPGM21}, as well as devising new methods to include background knowledge such as, in our our case, technical terminology and domain adaptation capabilities \cite{yu-etal-2021-adaptsum}, into multilingual pre-trained models.

Our work crucially builds upon recent advances in multilingual pre-trained models \cite{liu_multilingual_2020} and cross-lingual summarization \cite{ladhak-etal-2020-wikilingua}, and investigates how these methodologies can be applied for multilingual scholarly document processing. The application of NLP techniques for mining scientific papers has been primarily focused on the English language: with this work we want to put forward the vision of enabling scholarly document processing for a wider range of languages, ideally including both resource-rich and resource-poor languages in the longer term. Our vision of `Scholarly Document Processing for all languages' is in line with current trends in NLP (cf., e.g., \cite{ruder-etal-2021-xtreme} and \cite{adelani2021masakhaner}, \textit{inter alia}): while our initial effort concentrated here on fairly resource-rich languages, in future work we plan to focus specifically on resource-poor languages where multilingual NLP can and is indeed expected to make a difference in enabling wider (and consequently more diverse and  fairer \cite{joshi-etal-2020-state}) accessibility to scholarly resources.

\section*{Downloads}
The X-SCITLDR corpus and the code used in our CL-TLDR experiments is available under an open license at \url{https://github.com/sobamchan/xscitldr}.

\begin{acks}
The work presented in this paper is funded by the German Research Foundation (DFG) under the VADIS (PO 1900/5-1; EC 477/7-1) and JOIN-T2 (PO 1900/1-2) projects. We thank Ines Rehbein and the three anonymous reviewers for their helpful comments.
\end{acks}

\bibliographystyle{ACM-Reference-Format}
\bibliography{references,extras}

\end{document}